\newif\ifRAL
\RALtrue 	% RAL PAPER
\newif\ifTR
\TRfalse 	% Tech report not in

\newif\ifPrePrint
\PrePrinttrue
\ifRAL
  \documentclass[letterpaper, 10 pt, journal, twoside]{IEEEtran}
\else
  \documentclass[letterpaper, 10 pt, conference]{ieeeconf}
\fi

\IEEEoverridecommandlockouts

\ifRAL
\else
	\def\baselinestretch{0.90}
\fi

\makeatletter

\let\proof\@undefined
\let\endproof\@undefined
\makeatother

\usepackage{amsmath} % assumes amsmath package installed
\usepackage{amssymb}  % assumes amsmath package installed
\usepackage{amsthm}
\usepackage{amsfonts}
\usepackage{mathtools}
\usepackage{multirow}
\usepackage{array}

\usepackage{algpseudocode}
\usepackage{algorithm}

\usepackage{tikz}
\usepackage{tikzscale}
\usetikzlibrary{shapes,fit,arrows,positioning,calc}

\usepackage{bm}
\providecommand{\bm}{\pmb}

\newtheorem{prop}{Proposition}

\theoremstyle{definition}

\newtheorem{problem}{Problem}

\theoremstyle{remark}

\newtheorem{ass}{Assumption}

\usepackage{color}
\usepackage{graphicx}
\usepackage{caption}

\usepackage[bookmarks]{hyperref} %pdf with links and toc on the left
\hypersetup{
    colorlinks,
    citecolor=black,
    filecolor=black,
    linkcolor=black,
    urlcolor=black,
    pdfauthor={},
    pdfsubject={},
    pdftitle={}
}

\usepackage{url}

\usepackage[disable]{todonotes}

\graphicspath{{images/}}

\usepackage{graphicx}
\usepackage{color}

\usepackage{cite}

\usepackage{caption}

\usepackage{multirow}

\usepackage{paralist}
\usepackage{placeins}
\usepackage{subcaption}

\usepackage{color}
\newcommand\red[1]{{\textcolor{red}{#1}}}

\ifPrePrint
\newcommand{\rev}[1]{#1}
\else
\newcommand{\rev}[1]{#1}
\fi

\definecolor{myred}{rgb}{0.8500, 0.3250, 0.0980}
\definecolor{myblue}{rgb}{0    0.4470    0.7410}
\definecolor{mygreen}{rgb}{0.4660, 0.6740, 0.1880}
\definecolor{mypurple}{RGB}{139,0,139}

\usepackage[normalem]{ulem}

\newcommand{\fig}{Fig.~}	% figure ref
\DeclareMathOperator*{\argmin}{arg\,min}
\newcommand{\ie}{{\it i.e.},~}

\newcommand{\norm}[1]{\lVert#1\rVert}
\newcommand{\vect}[1]{\bm{#1}}		% vectors
\newcommand{\nR}[1]{\mathbb{R}^{#1}}		% real number
\newcommand{\xd}[1]{\vect{D}_{{#1}}}		
\newcommand{\xs}[1]{\vect{S}_{{#1}}}

\newcommand{\us}[1]{\vect{u}_{s_{#1}}}	
\newcommand{\ud}[1]{\vect{u}_{d_{#1}}}
\newcommand{\as}[1]{\vect{a}_{s_{#1}}}

\newcommand{\traj}{\bar{\vect{S}}}
\newcommand{\Vtraj}{\dot{\bar{\vect{S}}}}
\newcommand{\Atraj}{\ddot{\bar{\vect{S}}}}

\newcommand{\measj}[1]{\hat{\vect{y}}_j}

\newcommand{\pathVar}[1]{\hat{\vect{S}}_{#1}}

\ifRAL % RAL
			\author{Mahmoud Hamandi$^{1}$, Farshad Khorrami$^{1,2}$, Anthony Tzes$^{1}$%=======
	  \thanks{$^1$Center for Artificial Intelligence and Robotics, New York
University Abu Dhabi, Saadiyat Island, 129188, Abu Dhabi, UAE {\tt\footnotesize mahmoud.hamandi@nyu.edu, anthony.tzes@nyu.edu}}
\thanks{$^2$New York University, Brooklyn, NY, 11201, United States of America {\tt\footnotesize khorrami@nyu.edu}}
	    \thanks{\hspace{-1.0em} 
    This work has been supported by the NYUAD Center for Artificial Intelligence and Robotics, funded by Tamkeen under the NYUAD Research Institute Award CG010.}
	}

\else % ICRA
				\author{Mahmoud Hamandi$^{1}$, Farshad Khorrami$^{1,2}$, Anthony Tzes$^{1}$ 
	\thanks{$^1$Center for Artificial Intelligence and Robotics, New York
University Abu Dhabi, Saadiyat Island, 129188, Abu Dhabi, UAE {\tt\footnotesize mahmoud.hamandi@nyu.edu, anthony.tzes@nyu.edu}}
\thanks{$^2$New York University, Brooklyn, NY, 11201, United States of America {\tt\footnotesize khorrami@nyu.edu}}
	    \thanks{\hspace{-1.0em} 
    This work has been supported by the NYUAD Center for Artificial Intelligence and Robotics, funded by Tamkeen under the NYUAD Research Institute Award CG010.}
	}
\fi

\ifRAL % RAL
	\title{Robotic Shepherding in Cluttered and Unknown \\Environments using Control Barrier Functions}

\else % ICRA
	\title{\bf \textbf{Robotic Shepherding in Cluttered and Unknown \\Environments using Control Barrier Functions}}
\fi

\ifPrePrint
\else
\fi	
\begin{document}
%%%%%%%%%%%%%%%%%%%%%%%%%%%%%%%%%%%%%%%%%%%%%%%%%%%%%%%%%%%%%%%%%%%%%%

\maketitle
%%%%%%%%%%%%%%%%%%%%%%%%%%%%%%%%%%%%%%%%%%%%%%%%%%%%%%%%%%%%%%%%%%%%%%
\begin{abstract}
 
This paper introduces a novel control methodology designed to guide a collective of robotic-sheep in a cluttered and unknown environment using robotic-dogs. The dog-agents continuously scan the environment and compute a safe trajectory to guide the sheep to their final destination. The proposed optimization-based controller guarantees that the sheep reside within a desired distance from the reference trajectory through the use of Control Barrier Functions (CBF). Additional CBF constraints are employed simultaneously to ensure inter-agent and obstacle collision avoidance. The efficacy of the proposed approach is rigorously tested in simulation, which demonstrates the successful herding of the robotic-sheep within complex and cluttered environments.
\end{abstract}
%%%%%%%%%%%%%%%%%%%%%%%%%%%%%%%%%%%%%%%%%%%%%%%%%%%%%%%%%%%%%%%%%%%%%%
% \\\\\\\ Keywords \\\\\\
\ifRAL % RAL
	% Keywords appear just beneath the abstract. Use only for final RAL version.
\else % ICRA
	{} % nothing
\fi
%%%%%%%%%%%%%%%%%%%%%%%%%%%%%%%%%%%%%%%%%%%%%%%%%%%%%%%%%%%%%%%%%%%%%%
%\section*{Supplementary Material}
%%%%%%%%%%%%%%%%%%%%%%%%%%%%%%%%%%%%%%%%%%%%%%%%%%%%%%%%%%%%%%%%%%%%%%
%Open source design: \href{https://mrtbrnz.github.io/RoBust/}{\darkpurple{https://mrtbrnz.github.io/RoBust/}}.

%%%%%%%%%%%%%%%%%%%%%%%%%%%%%%%%%%%%%%%%%%%%%%%%%%%%%%%%%%%%%%%%%%%%%%
\section{INTRODUCTION}\label{sec:intro}
% Decomment only if accepted to RAL ----
\ifRAL
  \IEEEPARstart{S}{warm}
\else
   Swarm
\fi
technology has grown drastically in several applications over the last decade. The ability to achieve complex collective behaviors with small and simple robots have made the use of swarms interesting for many applications such as surveillance, attack and defense \cite{lachow2017upside,yin2021research}.  

%As swarms of Unmanned Aerial Vehicles (UAV)s are being allocated more often in military applications to overwhelm opponents, and thus effectively attack key targets, it becomes imperative to develop counter measures. In this perspective, t
%This paper suggests a swarm application, using defender robots to counter a group of attacker ones to alter their motion and escort them to a safe location. This task requires multiple algorithms for the defender swarm to \begin{inparaenum}
%\item entrap the attackers, 
%\item %hunt the attackers or 
%escort the attackers towards their final location
%\item and finally, entrap the attackers again in the final location
%\end{inparaenum} \cite{long2020comprehensive,antonelli2008entrapment,zhu2015multi}. %In this paper we suggest a novel algorithm that can achieve the different required tasks concurrently.

%It is noted that, in many papers in the literature, attacker and defender swarms are referred to as herds of sheep and herding dogs. This nomenclature is historically attributed due to the similarities between this problem and sheep shepherding \cite{long2020comprehensive}. Without loss of generality, in what follows we will also refer to attackers as sheep and to defenders as dogs.

\rev{This paper introduces a swarm application using two groups of swarms: robot-dogs (dogs) and  robot-sheep (sheep). Historically, this naming is due to the resemblance between the presented problem to the sheep herding by shepherd dogs. In the presented application, dogs are required to escort the sheep herd to a final destination along a safe path, while exploring and maneuvering around the unknown environment. Such scenario can be used for example to counter a group of drones and escort them to a location to neutralize their danger. Conversely, this scenario also arises in civilian applications, where a few intelligent robots herd a larger group of simple robots while the latter collect data from the environment.}

While the motion of the sheep herd is usually assumed to follow the dynamics %framed by Reynolds in their seminal work 
from~\cite{reynolds1987flocks}, many strategies for herder dogs to entrap and escort their adversary can be traced in the literature \cite{long2020comprehensive, antonelli2008entrapment, zhu2015multi}. 
The most common approach is to form a cage around the sheep, and slide the center of the cage along a trajectory leading to the final goal \cite{van2023steering}. As the sheep are assumed to be repelled by the dogs, a well placed cage can entrap the herd inside the cage.
Among the different caging algorithms, a very popular approach is the formation of a circle around the herd. \cite{bacon2012swarm}~forms a circle around a single sheep to steer it towards a final goal. \cite{pierson2015bio,pierson2017controlling} cage a sheep herd with a smaller group of dogs forming a circle around the sheep. As the cage slides along the desired trajectory, the dogs steer the sheep  and change their direction of motion as needed. While the circle formation is an interesting approach, it requires dogs to stay in formation to avoid sheep from escaping, and thus cannot easily accommodate complex environments%, or peculiar herding trajectories
.

To alleviate these limitations, \rev{the authors in~}\cite{chipade2021multiagent} connect dogs with a string net, which prevents the sheep from escaping the cage once trapped. Their innovative approach enabled them to shepherd multiple groups of sheep around obstacles. However, their approach requires the physical interaction between the net and the sheep, while it is more practical to manipulate sheep movements solely through non-physical forces. Sebastian et al. \cite{sebastian2021multi,sebastian2022adaptive} propose an interesting approach for the escorting problem, where they suggest that a group of dogs can "implicitly control" a group of sheep to follow a desired trajectory. While their method is promising, it requires the design of a trajectory for each sheep, and the active tracking of the corresponding trajectories; on the other hand, actively caging the sheep along a single trajectory can result in smoother and more feasible actions for the dogs.

Recently, in \cite{zhang2024distributed}, the authors proposed a distributed method where dogs ``push" sheep toward a desired goal through a distributed control approach. While this approach eliminates the necessity for inter-agent communication, it doesn't address herding in cluttered environments where complex trajectories are required to guide the sheep towards the desired goal. Closely related to this work, the authors in \cite{grover2022noncooperative,%mohanty2023distributed
mohanty2022distributed} showed a herding approach, where a group of dogs enforce a group of sheep to stay inside %(or outside) of 
a desired protected region. Their work focuses on a region modeled as a fixed circle and provides guarantees of convergence in the case of equal number of dogs and sheep, while dismissing the speed limit constraints of the different agents.  

The aim of this paper is to steer a noncooperative herd of sheep repelled by a controllable group of dogs to a goal location, while the latter agents explore the unknown and cluttered environment to find a safe path to the goal destination. The assumption throughout the paper is that the sheep will follow the herd dynamics suggested in~\cite{reynolds1987flocks}, however, their actions cannot be directly controlled. The shepherding objective is then to entrap the sheep herd within a protected area only through the control of the dogs' actions, as shown in Fig.~\ref{fig:overview}. The protected area is assumed to translate from the initial position of the sheep to the final position designated by the dogs. 
The protected area is enforced by the dogs' actions through the use of Control Barrier Functions (CBF)s~\cite{nagumo1942lage,prajna2004safety,wieland2007constructive}. While the CBF framework guarantees the caging of the sheep, it allows additionally the incorporation of different environment requirements, such as obstacle avoidance and inter-agent collision avoidance.

The contributions of this paper are as follows:
\begin{itemize}
    \item the introduction of a CBF-based herding controller, that allows dogs to cage the noncooperative sheep at a desired location and coerce them to follow a desired trajectory. \item the computation of a safe herding trajectory that steers the sheep towards the desired goal, while navigating and exploring obstacles in a cluttered and unknown environment; this environment is collectively identified by the dogs' distance sensors (LiDARs).
\end{itemize}

 The advantage of the proposed approach over others is that it does not require the hand crafting of a method to entrap the sheep or herd them along the desired trajectory, but rather only requires the definition of the desired sheep-behavior. While the sheep velocities cannot be directly controlled, the desired behavior can be guaranteed through the CBF-based controller determining the desired dog velocities. Correspondingly, the proposed method takes advantage of the guarantees provided by CBFs, which were demonstrated in problems such as area protection~\cite{grover2022noncooperative,mohanty2022distributed}, avoiding inter-agent collisions~\cite{sebastian2022adaptive,gonccalves2024safe}, swarm flocking~\cite{ibuki2020optimization}, swarm control in dynamic environments~\cite{10.5555/3635637.3663150}, obstacle avoidance~\cite{dai2023safe}, in addition to a plethora of applications that extend beyond the field of this paper~\cite{ames2019control,gonccalves2024control}.

The remainder of this paper is organized as follows:
Section~\ref{sec:modeling} introduces the models for the system dynamics. Problem statement and the assumptions made about the agents and the environment are given in Section~\ref{sec:prob_stat_assum}. %Section~\ref{sec:cbf} provides a quick overview of Control Barrier Functions, and 
Section~\ref{sec:frontier_exploration} details the frontier exploration and trajectory computation. 
Section~\ref{sec:ctrl} introduces the CBF herding algorithm in the presence of obstacles. Finally, Section~\ref{sec:results} demonstrates the efficacy of the proposed algorithm by simulation studies.
%%%%%%%%%%%%%%%%%%%%%%%%%%%%%%%%%%%%%%%%%%%%%%%%%%%%%%%%%%%%%%%%%%%%%%

%%%%%%%%%%%%%%%%%%%%%%%%%%%%%%%%%%%%%%%%%%%%%%%%%%%%%%%%%%%%%%%%%%%%%%
\section{Modeling}\label{sec:modeling}
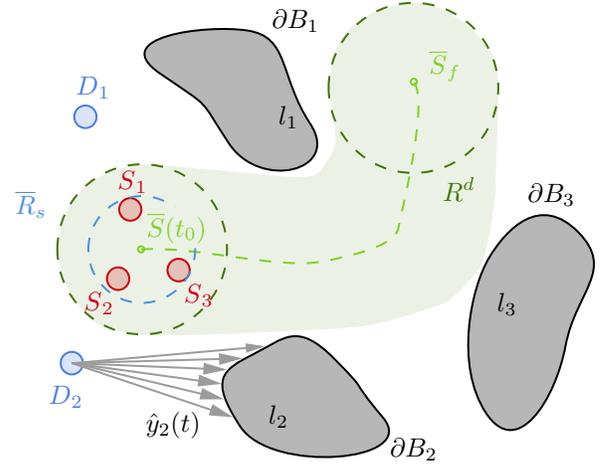
\begin{figure}[htbp]
    \centering

\tikzset{every picture/.style={line width=0.75pt}} %set default line width to 0.75pt        

\begin{tikzpicture}[x=0.75pt,y=0.75pt,yscale=-1,xscale=1]

%Shape: Polygon Curved [id:ds6257916083535249] 
\draw  [draw opacity=0][fill={rgb, 255:red, 65; green, 117; blue, 5 }  ,fill opacity=0.1 ] (100.67,135.67) .. controls (105.26,133.37) and (113.9,131.76) .. (118.8,131.18) .. controls (123.7,130.61) and (184.2,137.27) .. (196.75,137.63) .. controls (209.3,137.98) and (208.71,134.13) .. (213,128.13) .. controls (217.29,122.13) and (216.57,118.71) .. (215.43,108.71) .. controls (214.29,98.71) and (211.57,99.14) .. (214.57,80.14) .. controls (217.57,61.14) and (234.75,52.13) .. (253.71,49.86) .. controls (272.68,47.59) and (294.48,69.52) .. (297.14,79.86) .. controls (299.81,90.19) and (297.25,124) .. (297.14,139) .. controls (297.03,154) and (292.35,175.43) .. (290.57,180.43) .. controls (288.8,185.43) and (276.59,198.06) .. (273.29,200.18) .. controls (270,202.29) and (243.4,210.6) .. (236.29,211.86) .. controls (229.17,213.11) and (224.47,213.82) .. (204.81,214.78) .. controls (185.15,215.73) and (141.75,218.13) .. (122,217.33) .. controls (102.25,216.54) and (93.89,211.17) .. (90.67,207) .. controls (87.44,202.83) and (81.95,197.74) .. (78.67,187) .. controls (75.39,176.26) and (77.05,169.56) .. (77.33,165) .. controls (77.61,160.44) and (80.46,153.96) .. (86.38,146.75) .. controls (92.29,139.54) and (96.08,137.96) .. (100.67,135.67) -- cycle ;
%Shape: Ellipse [id:dp9407342082440884] 
\draw  [color={rgb, 255:red, 74; green, 125; blue, 226 }  ,draw opacity=1 ][fill={rgb, 255:red, 74; green, 125; blue, 226 }  ,fill opacity=0.2 ] (84.55,107.22) .. controls (84.55,104.08) and (87.06,101.53) .. (90.16,101.53) .. controls (93.26,101.53) and (95.77,104.08) .. (95.77,107.22) .. controls (95.77,110.36) and (93.26,112.91) .. (90.16,112.91) .. controls (87.06,112.91) and (84.55,110.36) .. (84.55,107.22) -- cycle ;
%Shape: Ellipse [id:dp35931372035246745] 
\draw  [color={rgb, 255:red, 208; green, 2; blue, 2 }  ,draw opacity=1 ][fill={rgb, 255:red, 208; green, 2; blue, 2 }  ,fill opacity=0.2 ] (101.06,188.92) .. controls (101.06,185.78) and (103.57,183.23) .. (106.67,183.23) .. controls (109.76,183.23) and (112.27,185.78) .. (112.27,188.92) .. controls (112.27,192.06) and (109.76,194.61) .. (106.67,194.61) .. controls (103.57,194.61) and (101.06,192.06) .. (101.06,188.92) -- cycle ;
%Shape: Ellipse [id:dp8966102966225831] 
\draw  [color={rgb, 255:red, 65; green, 117; blue, 5 }  ,draw opacity=1 ][dash pattern={on 4.5pt off 4.5pt}] (76.02,174.12) .. controls (76.02,150.41) and (95.18,131.18) .. (118.8,131.18) .. controls (142.43,131.18) and (161.58,150.41) .. (161.58,174.12) .. controls (161.58,197.83) and (142.43,217.05) .. (118.8,217.05) .. controls (95.18,217.05) and (76.02,197.83) .. (76.02,174.12) -- cycle ;
%Shape: Ellipse [id:dp04781480985993691] 
\draw  [color={rgb, 255:red, 126; green, 211; blue, 33 }  ,draw opacity=1 ] (116.69,174.12) .. controls (116.69,173.24) and (117.39,172.54) .. (118.25,172.54) .. controls (119.1,172.54) and (119.8,173.24) .. (119.8,174.12) .. controls (119.8,174.99) and (119.1,175.7) .. (118.25,175.7) .. controls (117.39,175.7) and (116.69,174.99) .. (116.69,174.12) -- cycle ;
%Shape: Polygon Curved [id:ds4903517459498661] 
\draw  [fill={rgb, 255:red, 183; green, 183; blue, 183 }  ,fill opacity=1 ] (174.78,227.09) .. controls (190.58,219.08) and (196.89,210.26) .. (215.84,229.49) .. controls (234.79,248.72) and (221.63,236.84) .. (240.32,259.14) .. controls (259.01,281.44) and (184.1,287.66) .. (168.31,263.62) .. controls (152.52,239.59) and (158.99,235.1) .. (174.78,227.09) -- cycle ;
%Shape: Ellipse [id:dp48770774033701914] 
\draw  [color={rgb, 255:red, 208; green, 2; blue, 2 }  ,draw opacity=1 ][fill={rgb, 255:red, 208; green, 2; blue, 2 }  ,fill opacity=0.2 ] (131.57,184.59) .. controls (131.57,181.45) and (134.08,178.9) .. (137.17,178.9) .. controls (140.27,178.9) and (142.78,181.45) .. (142.78,184.59) .. controls (142.78,187.73) and (140.27,190.28) .. (137.17,190.28) .. controls (134.08,190.28) and (131.57,187.73) .. (131.57,184.59) -- cycle ;
%Shape: Ellipse [id:dp2525110551965426] 
\draw  [color={rgb, 255:red, 208; green, 2; blue, 2 }  ,draw opacity=1 ][fill={rgb, 255:red, 208; green, 2; blue, 2 }  ,fill opacity=0.2 ] (107.19,153.97) .. controls (107.19,150.83) and (109.7,148.28) .. (112.8,148.28) .. controls (115.89,148.28) and (118.4,150.83) .. (118.4,153.97) .. controls (118.4,157.11) and (115.89,159.66) .. (112.8,159.66) .. controls (109.7,159.66) and (107.19,157.11) .. (107.19,153.97) -- cycle ;
%Shape: Polygon Curved [id:ds40711430659257597] 
\draw  [fill={rgb, 255:red, 183; green, 183; blue, 183 }  ,fill opacity=1 ] (127.73,85.32) .. controls (100.89,66.89) and (150.63,60.48) .. (171.95,63.68) .. controls (193.27,66.89) and (187.75,86.68) .. (203.54,110.71) .. controls (219.33,134.75) and (178.27,144.6) .. (162.48,120.57) .. controls (146.69,96.53) and (154.58,103.74) .. (127.73,85.32) -- cycle ;
%Shape: Polygon Curved [id:ds24439431263545197] 
\draw  [fill={rgb, 255:red, 183; green, 183; blue, 183 }  ,fill opacity=1 ] (293.64,178.57) .. controls (302.75,163.27) and (319.72,146.16) .. (338.68,165.39) .. controls (357.63,184.62) and (337.49,195.43) .. (332.36,222.67) .. controls (327.23,249.91) and (307.96,280.72) .. (292.17,256.68) .. controls (276.37,232.65) and (284.53,193.86) .. (293.64,178.57) -- cycle ;
%Shape: Ellipse [id:dp47650220213861516] 
\draw  [color={rgb, 255:red, 74; green, 125; blue, 226 }  ,draw opacity=1 ][fill={rgb, 255:red, 74; green, 125; blue, 226 }  ,fill opacity=0.2 ] (77.71,231.56) .. controls (77.71,228.42) and (80.22,225.87) .. (83.31,225.87) .. controls (86.41,225.87) and (88.92,228.42) .. (88.92,231.56) .. controls (88.92,234.7) and (86.41,237.25) .. (83.31,237.25) .. controls (80.22,237.25) and (77.71,234.7) .. (77.71,231.56) -- cycle ;
%Curve Lines [id:da841985412350972] 
\draw [color={rgb, 255:red, 126; green, 211; blue, 33 }  ,draw opacity=1 ] [dash pattern={on 4.5pt off 4.5pt}]  (118.25,174.12) .. controls (146.76,172.33) and (187.89,184.66) .. (210.57,180.14) .. controls (233.25,175.63) and (246.67,170.55) .. (248,163.38) .. controls (249.33,156.2) and (263.7,99.75) .. (255.8,89.6) ;
%Shape: Ellipse [id:dp20183803756228813] 
\draw  [color={rgb, 255:red, 126; green, 211; blue, 33 }  ,draw opacity=1 ] (254.24,89.6) .. controls (254.24,88.73) and (254.94,88.02) .. (255.8,88.02) .. controls (256.66,88.02) and (257.36,88.73) .. (257.36,89.6) .. controls (257.36,90.47) and (256.66,91.18) .. (255.8,91.18) .. controls (254.94,91.18) and (254.24,90.47) .. (254.24,89.6) -- cycle ;
%Straight Lines [id:da20534596506371616] 
\draw [color={rgb, 255:red, 155; green, 155; blue, 155 }  ,draw opacity=1 ]   (83.31,231.56) -- (167.54,229.21) ;
\draw [shift={(169.54,229.15)}, rotate = 178.4] [fill={rgb, 255:red, 155; green, 155; blue, 155 }  ,fill opacity=1 ][line width=0.08]  [draw opacity=0] (12,-3) -- (0,0) -- (12,3) -- cycle    ;
%Straight Lines [id:da48447650449179935] 
\draw [color={rgb, 255:red, 155; green, 155; blue, 155 }  ,draw opacity=1 ]   (83.31,231.56) -- (159.69,234.76) ;
\draw [shift={(161.69,234.85)}, rotate = 182.4] [fill={rgb, 255:red, 155; green, 155; blue, 155 }  ,fill opacity=1 ][line width=0.08]  [draw opacity=0] (12,-3) -- (0,0) -- (12,3) -- cycle    ;
%Straight Lines [id:da06669120188896849] 
\draw [color={rgb, 255:red, 155; green, 155; blue, 155 }  ,draw opacity=1 ]   (83.31,231.56) -- (157.02,241.73) ;
\draw [shift={(159,242)}, rotate = 187.85] [fill={rgb, 255:red, 155; green, 155; blue, 155 }  ,fill opacity=1 ][line width=0.08]  [draw opacity=0] (12,-3) -- (0,0) -- (12,3) -- cycle    ;
%Straight Lines [id:da18862108763508756] 
\draw [color={rgb, 255:red, 155; green, 155; blue, 155 }  ,draw opacity=1 ]   (84.31,231.56) -- (178.64,223.27) ;
\draw [shift={(180.64,223.09)}, rotate = 174.97] [fill={rgb, 255:red, 155; green, 155; blue, 155 }  ,fill opacity=1 ][line width=0.08]  [draw opacity=0] (9.6,-2.4) -- (0,0) -- (9.6,2.4) -- cycle    ;
%Straight Lines [id:da034960170118004674] 
\draw [color={rgb, 255:red, 155; green, 155; blue, 155 }  ,draw opacity=1 ]   (84.31,231.56) -- (158.06,249.3) ;
\draw [shift={(160,249.77)}, rotate = 193.53] [fill={rgb, 255:red, 155; green, 155; blue, 155 }  ,fill opacity=1 ][line width=0.08]  [draw opacity=0] (12,-3) -- (0,0) -- (12,3) -- cycle    ;
%Straight Lines [id:da6149738607140305] 
\draw [color={rgb, 255:red, 155; green, 155; blue, 155 }  ,draw opacity=1 ]   (83.31,231.56) -- (162.77,258.23) ;
\draw [shift={(164.67,258.87)}, rotate = 198.55] [fill={rgb, 255:red, 155; green, 155; blue, 155 }  ,fill opacity=1 ][line width=0.08]  [draw opacity=0] (12,-3) -- (0,0) -- (12,3) -- cycle    ;
%Shape: Ellipse [id:dp09509183461142823] 
\draw  [color={rgb, 255:red, 74; green, 144; blue, 226 }  ,draw opacity=1 ][dash pattern={on 4.5pt off 4.5pt}] (91.64,174.21) .. controls (91.64,159.32) and (103.7,147.25) .. (118.57,147.25) .. controls (133.44,147.25) and (145.5,159.32) .. (145.5,174.21) .. controls (145.5,189.09) and (133.44,201.16) .. (118.57,201.16) .. controls (103.7,201.16) and (91.64,189.09) .. (91.64,174.21) -- cycle ;
%Shape: Ellipse [id:dp6880642778818664] 
\draw  [color={rgb, 255:red, 65; green, 117; blue, 5 }  ,draw opacity=1 ][dash pattern={on 4.5pt off 4.5pt}] (212.83,92.64) .. controls (212.83,68.93) and (231.99,49.71) .. (255.61,49.71) .. controls (279.24,49.71) and (298.39,68.93) .. (298.39,92.64) .. controls (298.39,116.35) and (279.24,135.58) .. (255.61,135.58) .. controls (231.99,135.58) and (212.83,116.35) .. (212.83,92.64) -- cycle ;
%Shape: Rectangle [id:dp6950942099921522] 
\draw  [draw opacity=0][fill={rgb, 255:red, 255; green, 255; blue, 255 }  ,fill opacity=1 ] (121.47,152.82) -- (152.06,152.82) -- (152.06,171.88) -- (121.47,171.88) -- cycle ;
%Shape: Rectangle [id:dp8884341517434287] 
\draw  [draw opacity=0][fill={rgb, 255:red, 65; green, 117; blue, 5 }  ,fill opacity=0.1 ] (121.47,152.82) -- (152.06,152.82) -- (152.06,171.88) -- (121.47,171.88) -- cycle ;

% Text Node
\draw (119.77,154.26) node [anchor=north west][inner sep=0.75pt]   [align=left] {$\displaystyle \textcolor[rgb]{0.49,0.83,0.13}{\overline{S}}\textcolor[rgb]{0.49,0.83,0.13}{(}\textcolor[rgb]{0.49,0.83,0.13}{t}\textcolor[rgb]{0.49,0.83,0.13}{_{0}}\textcolor[rgb]{0.49,0.83,0.13}{)}$};
% Text Node
\draw (268.94,136.17) node [anchor=north west][inner sep=0.75pt]   [align=left] {$\displaystyle \textcolor[rgb]{0.25,0.46,0.02}{R^{d}}$};
% Text Node
\draw (262.12,72.31) node [anchor=north west][inner sep=0.75pt]   [align=left] {$\displaystyle \textcolor[rgb]{0.49,0.83,0.13}{\overline{S}_{f}}$};
% Text Node
\draw (83.82,85.42) node [anchor=north west][inner sep=0.75pt]   [align=left] {$\displaystyle \textcolor[rgb]{0.29,0.49,0.89}{D}\textcolor[rgb]{0.29,0.49,0.89}{_{1}}$};
% Text Node
\draw (70.13,241.41) node [anchor=north west][inner sep=0.75pt]   [align=left] {$\displaystyle \textcolor[rgb]{0.29,0.49,0.89}{D}\textcolor[rgb]{0.29,0.49,0.89}{_{2}}$};
% Text Node
\draw (105.03,132.84) node [anchor=north west][inner sep=0.75pt]   [align=left] {$\displaystyle \textcolor[rgb]{0.82,0.01,0.11}{S}\textcolor[rgb]{0.82,0.01,0.11}{_{1}}$};
% Text Node
\draw (88.11,194.08) node [anchor=north west][inner sep=0.75pt]   [align=left] {$\displaystyle \textcolor[rgb]{0.82,0.01,0.11}{S}\textcolor[rgb]{0.82,0.01,0.11}{_{2}}$};
% Text Node
\draw (138.44,190.33) node [anchor=north west][inner sep=0.75pt]   [align=left] {$\displaystyle \textcolor[rgb]{0.82,0.01,0.11}{S}\textcolor[rgb]{0.82,0.01,0.11}{_{3}}$};
% Text Node
\draw (186.19,100.09) node [anchor=north west][inner sep=0.75pt]   [align=left] {$\displaystyle l_{1}$};
% Text Node
\draw (181.1,251.39) node [anchor=north west][inner sep=0.75pt]   [align=left] {$\displaystyle l_{2}$};
% Text Node
\draw (296.44,194.54) node [anchor=north west][inner sep=0.75pt]   [align=left] {$\displaystyle l_{3}$};
% Text Node
\draw (182.7,51.65) node [anchor=north west][inner sep=0.75pt]   [align=left] {$\displaystyle \partial B_{1}$};
% Text Node
\draw (242.27,268) node [anchor=north west][inner sep=0.75pt]   [align=left] {$\displaystyle \partial B_{2}$};
% Text Node
\draw (311.73,139.45) node [anchor=north west][inner sep=0.75pt]   [align=left] {$\displaystyle \partial B_{3}$};
% Text Node
\draw (119.13,254.41) node [anchor=north west][inner sep=0.75pt]   [align=left] {$\displaystyle \hat{y}_{2}( t)$};
% Text Node
\draw (53.25,143.13) node [anchor=north west][inner sep=0.75pt]   [align=left] {$\displaystyle \textcolor[rgb]{0.29,0.56,0.89}{\overline{R}_{s}}$};

\end{tikzpicture}
    \caption{Overview of the sheep herding algorithm.} 
    \label{fig:overview}
\end{figure}

%%%%%%%%%%%%%%%%%%%%%%%%%%%%%%%%%%%%%%%%%%%%%%%%%%%%%%%%%%%%%%%%%%%%%%
The objective of this paper is to control a swarm of $m$ dogs herding a swarm of $n$ sheep in a 2D-planar unknown environment, as shown in Fig.~\ref{fig:overview}.

Denote the group of sheep and dogs  as $\mathcal{S}$ and $\mathcal{D}$, respectively, \rev{where each of the two sets contains the indices of the different agents in the corresponding set}. Denote  $\xs{i} \in \nR{2}$  and $\xd{j} \in \nR{2}$ as the corresponding position of the $i$-th sheep and the ${j\text{-th}}$ dog, and  $\us{i}$ and $\ud{j}$ the corresponding velocities of the agents modeled as single integrator point-masses. For compactness, define ${\xs{} = [\xs{1}^\top \cdots \xs{n}^\top]^\top \in \nR{2 n}}$ and 
${\xd{} = [\xd{1}^\top \cdots \xd{m}^\top]^\top \in \nR{2 m}}$ as the positions of all sheep and all dogs, and  $\us{} \in \nR{2 n}$ and $\ud{} \in \nR{2 m}$ their corresponding velocities. It is assumed that all agents in the environment have velocity limits, i.e., ${-\bar{v} \leq\rev{ u_{s_i}} \leq \bar{v} \;\; \forall i \in \mathcal{S}}$ and %where $\bar{v}$ is the maximum possible sheep velocity in any direction;
$-\bar{u} \leq \rev{u_{d_j}} \leq \bar{u} \;\; \forall j \in \mathcal{D}$, where $\bar{v}~(\bar{u})$ is the maximum sheep~(dog) velocity in any direction.

Finally, we assume that dog velocities are directly controllable, while the sheep dynamics are governed by the inter-agent repulsion and attraction as follows
\vspace{-0.05cm}
\begin{eqnarray}\label{eq:sheep_dynamics}
\vect{f}_i &=& \dot{\xs{}}_{i} = \us{i} (\xs{}, \xd{}) = \bar{v} \;\text{tanh} (\us{i}^u(\xs{}, \xd{})/\bar{v}) \nonumber\\ 
    &=&
    \bar{v} \;\text{tanh} \left(\frac{k_s}{\bar{v}} \sum_{\rev{k}\in \mathcal{S}/i}\left(1-\frac{R_s^3}{\norm{\xs{\rev{k}} - \xs{i}}^3}\right)(\xs{\rev{k}} - \xs{i})  \right. \nonumber \\
    &&+\left.  \frac{k_d}{\bar{v}} \sum_{\rev{j}\in\mathcal{D}}\frac{\xs{i} - \xd{\rev{j}}}{\norm{\xs{i} - \xd{\rev{j}}}^3} \right)
\end{eqnarray}
where $k_s$, $k_d$ are gains related to inter-sheep attraction/repulsion and dog-sheep repulsion, and $R_s$ is the desired distance between sheep. Note that the `tanh' allows the smooth inclusion of the sheep speed limits into its dynamics, where $\us{i}^u(\xs{}, \xd{})$ denotes the non-constrained sheep velocity.

%In addition, it is noted that the sheep dynamics will encourage the sheep to be in formation throughout their trajectory, unless this formation was temporarily broken by dog agents. We denote by $\overline{R}_s$ as the radius of the minimum circumscribed circle that the sheep herd will reside in when in formation and undisturbed. This radius can be estimated from observing the behavior of the herd at steady state, \ie when $\us{i} = \vect{0} \,\, \forall i \in \mathcal{S}$ in the absence of dogs. Since dogs are always present in our environment, $\overline{R}_s$ is estimated from the sheep formation when sheep are stationary and the distance to dogs is large enough such that their effect is minimal on the corresponding formation.
It is noted that sheep tend to maintain formation unless disrupted by dogs. We denote $\overline{R}_s$ as the radius of the minimum circumscribed circle the undisturbed sheep herd forms, shown in Fig.~\ref{fig:overview}. This radius is estimated from the herd's steady state behavior, observed when sheep are stationary and dogs are distant enough to minimally affect the formation.

Note that $\us{i}$ is a function of dogs' positions, while the sheep accelerations $\as{i}$ are a function of the dogs' velocities as follows
\begin{align}
    \as{i} &= \sum_{k\in \mathcal{S}/i}\frac{\partial \vect{f}_i}{\partial \xs{k}} (\us{k} - \us{i})  + \sum_{j\in\mathcal{D}}\frac{\partial \vect{f}_i}{\partial \xd{j}} (\ud{j} - \us{i} ) %\label{eq:sheep_acc}
\end{align}
where 
\begin{align}
    \frac{\partial \vect{f}_i}{\partial \xs{\rev{k}}} &= s_i\frac{k_s R_s^3}{\bar{v}}\left(1-\frac{\vect{I}_2 d_{i,\rev{k}}^\top d_{i,\rev{k}} - 3d_{i,\rev{k}}d_{i,\rev{k}}^\top}{\norm{d_{i,\rev{k}}}^5} \right)\\
    \frac{\partial \vect{f}_i}{\partial \xd{\rev{j}}} &= -s_i\frac{k_d}{\bar{v}}\frac{\vect{I}_2 d_{i,\rev{j}}^\top d_{i,\rev{j}} - 3d_{i,\rev{j}}d_{i,\rev{j}}^\top}{\norm{d_{i,\rev{j}}}^5} 
\end{align}
 and $s_i = \text{sech}^2(\us{i}^u(\xs{}, \xd{})/\bar{v})$, $d_{i,\rev{k}} = \xs{\rev{k}} - \xs{i}$, ${d_{i,\rev{j}} = \xs{i} - \xd{\rev{j}}}$, and $\vect{I}_2 \in \nR{2}$ is the identity matrix.
 
%%%%%%%%%%%%%%%%%%%%%%%%%%%%%%%%%%%%%%%%%%%%%%%%%%%%%%%%%%%%%%%%%%%%%%
\section{Problem Statement and Assumptions}
\label{sec:prob_stat_assum}
%%%%%%%%%%%%%%%%%%%%%%%%%%%%%%%%%%%%%%%%%%%%%%%%%%%%%%%%%%%%%%%%%%%%%% 
The sheep and dogs are assumed to reside in an unknown environment where there exists an obstacle-free trajectory $\traj(t) \in \nR{2} \; \forall t \in [t_0 \; t_f] \in \mathcal{C}^2$ to be discovered from the sheep's initial position $\traj(t_0)$ to a final goal position $\traj_f = \traj(t_f)$, where $t_0$ and $t_f$ are the starting and ending time of the herding process. Note that the trajectory is not unique and depends on the path planning approach. The velocity and acceleration along the trajectory are denoted as ${\Vtraj(t) \in \nR{2}}$ and $\Atraj(t) \in \nR{2}$.
In addition, there exists a radius $R^d \geq \overline{R}_s \in \nR{} $ that specifies a circle centered at the trajectory $\traj(t)$.

Given the above, and following the discovery of the trajectory, the objective of this paper can be summarized as follows:
\begin{problem}
    Given $R^d, \, \traj(t) \; \forall t \in [ t_0\; t_f]$, find control inputs  $\ud{}(t) \;\forall t \in [ t_0\; t_f]$ such that
    \begin{align}
    & \xs{i} \in \mathcal{P}_i \;\; \forall i \in \mathcal{S}  \forall t \in [ t_0\; t_f] \mbox{, where} \nonumber \\
    & \mathcal{P}_i:= \{ \xs{i} \in \mathbb{R}^2| \norm{\xs{i} - \traj(t)} \leq R^d \;\; \forall t \in [ t_0\; t_f]\}.\label{eq:herding_protected_region}
    \end{align}
\end{problem}

Since it is desired to control the agents' motion, the following assumptions are made:
\begin{ass}Agents' motion assumptions:
     \begin{itemize}
         \item It is assumed that all sheep follow the dynamics given in \eqref{eq:sheep_dynamics}, and their motion can only be controlled via the placement of the dogs by controlling their velocities. In the absence of dogs, sheep will form a fixed topological cluster.
         \item It is assumed that the dogs' velocities $\ud{}$ are directly controlled since the dogs' dynamics correspond to an integrator.
     \end{itemize}
\end{ass}
The problem is to find the dogs' velocities that will achieve the desired sheep's motion through the environment, having the following assumptions:
\begin{ass}
    Environment assumptions:
    \begin{itemize}
        \item The environment contains $L$ obstacles, where each obstacle $l=1,\dots,L$ is denoted by its boundary $\partial B_l$.
        \item Nearby obstacles to all agents can be sensed by all agents for  collision avoidance, but only dog agents can achieve long range sensing required for mapping and path planning.
        \item The environment is unknown for all agents, and has to be discovered by the dogs to guide the sheep to the final goal.
        \item The closest distance between at least two pair of obstacles, denoted as $\underline{\partial l}$ satisfies ${2\overline{R}_s \leq \underline{\partial l} \;\; \forall t \in [t_0 \;t_f]}$; thus there exists a path in the environment where sheep can pass within their cluster.
    \end{itemize}
\end{ass}
%Similarly the trajectory's assumptions are:
%\begin{ass}
%    Trajectory assumptions:
%    \begin{itemize}
%        \item the trajectory is constructed such that $\traj(t)$ is at least $\overline{R}_s$ away from any obstacle $\forall t \in [t_0, t_f]$.
%        \item The desired trajectory is constructed using the skeleton~\cite{unlu2023uav} defined between the discovered obstacles.
%        \item The velocity and acceleration along the computed trajectory are bounded such that $\norm{\Vtraj}_\infty \leq \overline{\Vtraj}$ and $\norm{\Atraj}_\infty \leq \overline{\Atraj}$, where $\overline{\Vtraj}$ and $\overline{\Atraj}$ are chosen by the user to ensure the sheep's ability to follow the desired trajectory.
%    \end{itemize}
%\end{ass}

%This radius is specified by the user during the path planning process and should be compared to the minimum circle within which the sheep reside at each moment, ${R(t)=\max_{i} \norm{\xs{i}-\traj(t)}}$. \mh{this needs to be placed somewhere}

%%%%%%%%%%%%%%%%%%%%%%%%%%%%%%%%%%%%%%%%%%%%%%%%%%%%%%%%%%%%%%%%%%%%%%
%\section{Control Barrier Function: An Overview}\label{sec:cbf}
%%%%%%%%%%%%%%%%%%%%%%%%%%%%%%%%%%%%%%%%%%%%%%%%%%%%%%%%%%%%%%%%%%%%%% 
%%%%%%%%%%%%%%%%%%%%%%%%%%%%%%%%%%%%%%%%%%%%%%%%%%%%%%%%%%%%%%%%%%%%
\section{CBF Overview \& Frontier Exploration}\label{sec:frontier_exploration}
%%%%%%%%%%%%%%%%%%%%%%%%%%%%%%%%%%%%%%%%%%%%%%%%%%%%%%%%%%%%%%%%%%%%%%
\subsection{Control Barrier Functions}
The subsequent overview summarizes the necessary CBF concepts~\cite{nagumo1942lage,prajna2004safety,wieland2007constructive,dai2022learning}. CBFs ensure the safety of a dynamical system $\dot{x} = f(x,u) \in \nR{n}$, by enforcing the forward invariance of a safety set $\mathcal{P}$%, \ie if $x \in \mathcal{P}$ at time $t_0$, then $x \in \mathcal{P} \; \forall t \geq t_0$
.
Let %us define 
the set $\mathcal{P}$ be associated to the continuously differentiable function $h\;:\; X \in \nR{n} \rightarrow \nR{}$;
\begin{align}
    \mathcal{P} = \{x \in X \in \nR{n} \;:\; h(x) \geq 0 \}.
\end{align}
This set is forward invariant if there exists an extended class $\kappa_\infty$-function $\alpha$ such that 
\begin{align}
    \exists u \; \text{ such that} \;   \dot{h}(x,u) + \alpha(h) \geq 0 \; \forall x \in X
\end{align}
where a $\kappa_\infty$-function $\alpha$ is a strictly increasing function, $\exists a$ such that $\alpha(a) = \infty$, and $\lim_{r\rightarrow \infty} \alpha(r) = \infty$.

In the case where $\dot{h}$ depends explicitly on the control input $u$, the CBF is denoted as a CBF of degree-1, and the above constraint is enough to ensure the forward invariance of $\mathcal{P}$. However, if the control input's first explicit appearance is in the $k$-th derivative of $h$, then the CBF is denoted of ${\text{degree-}k}$. In this case, define the sequence of functions $\phi_i$ such that  $\phi_0 = h(x)$, and $\phi_i = \dot{\phi}_{i-1} + \alpha_i(\phi_{i-1})$, where the control input $u$ will appear explicitly in $\phi_k$. In this case, the forward invariance of $\mathcal{P}$ can be ensured through the following constraint:
\begin{align}
    \phi_{k}(x,u) = \dot{\phi}_{k-1}(x,u) + \alpha_i(\phi_{k-1}) \geq 0.
\end{align}

\subsection{Environment Mapping and Path Computation}\label{subsec:path}
Assume a centralized environment estimation and trajectory planning, using the collective LiDAR-scans of the different dogs.
Consider as $\mathcal{M}$ the centralized estimated occupancy grid map of the environment. While estimating the environment map, the positions of the different dog-agents are assumed to be known.
Let $\measj{j}(t)$ be the LiDAR measurements corresponding to the $j$th dog at its position $\xd{j}(t)$. The LiDAR scan consists of depth measures to obstacles in the environment around a dog's center. 
Following any LiDAR measurement from any dog, the map is updated \begin{align}
    \mathcal{M} \leftarrow \text{UpdateOccupancyMap}(\mathcal{M} | \measj{j}(t), \xd{j}(t)) .
\end{align}
The occupancy map is updated in a Simultaneous Localization and Mapping algorithm~\cite{7487258},
and augmented by marking any grid location that is closer than $\overline{R}_s$ to any scanned obstacle as non-traversable. This would ensure the constructed path is at least $\overline{R}_s$ away from obstacles.

Following the computation of the map, a path is computed from the desired trajectory position at the current time stamp $\traj(t_c)$ to the final desired position $\traj_f$ using the skeletal approach~\cite{unlu2023uav,chaussard2011robust}. The skeletal approach finds a path along the median line, situated equidistantly between obstacles. The computed path follows the skeleton of the map within the discovered area, and then follows a straight line from the last point along the skeleton in the discovered area to $\traj_f$. The computed path at the current time step $t_c$ is denoted by 
$\pathVar{}[k] \; \forall k \in [0,\ldots,K]$, with $\pathVar{}[0] = \traj(t_c)$, and $\pathVar{}[K] = \traj_f$, and $K$ being the length of the path. Note that each $\pathVar{}[k]$ corresponds to a grid location in the occupancy map, and that $\pathVar{}[k]$ and $\pathVar{}[k+1]$ are one grid location away from each other. %As such, following the corresponding path will ensure sheep will stay the furthest away from scanned obstacles.

A polynomial trajectory is fitted to the path while 
ensuring velocity and acceleration continuity at $t_c$ with the previously computed trajectory. It is common in the literature to segment the path into multiple segments, and to fit a polynomial to each~\cite{mellinger2011minimum}. This insures numerical stability of the solution and avoids oscillations. As such, we fit the trajectory to a segment of the computed path $\hat{S}[k] \in [0,\dots,H]$ such that $H\leq K$,
%up to a horizon point $H \leq K$, 
while optimizing the time $T$ to reach $\pathVar{}[H]$. The trajectory is computed while satisfying kinematic constraints on velocity and acceleration, ensuring that $\norm{\Vtraj(t)}_\infty \leq \overline{\Vtraj}$ and $\norm{\Atraj(t)}_\infty \leq \overline{\Atraj}$, where $\overline{\Vtraj}$ and $\overline{\Atraj}$ are chosen by the user to ensure the sheep's ability to follow the desired trajectory.
The computed trajectory is valid for $t\in [t_c,t_c+T]$, and a new trajectory is computed before or at the end of the current segment.
\section{Optimization Based Controller}\label{sec:ctrl}
%%%%%%%%%%%%%%%%%%%%%%%%%%%%%%%%%%%%%%%%%%%%%%%%%%%%%%%%%%%%%%%%%%%%%%
The objective of the controller is to allocate dog velocities to: 
\begin{inparaenum}
    \item coerce the sheep to stay within the protected region $\mathcal{P}_i$ as it translates along $\traj(t)$, 
    \item avoid obstacles, and
    \item avoid collision with other dogs.
\end{inparaenum}
These objectives are enforced as optimization constraints in a quadratic programming framework (QP) using a CBF to enforce each term.

%It is noted that each of the below CBF have a different degree; as such, each CBF will be derived $k$-times, such that the corresponding $k$-th derivative is an explicit expression of the dog velocities $\ud{}$. 
\subsection{Sheep herding CBF}
To enforce each sheep $i$ to stay within the protected area~\eqref{eq:herding_protected_region}, choose the following CBF with a safety radius $r$ that avoids sheep getting close to the edge of the protected region 
\begin{align}
    h^p_i = -\frac{1}{2}( \norm{\xs{i} - \traj(t)}^2 - (R^d-r)^2) \geq 0 .
\end{align}
The first and second derivatives of $h_i^p$ are
\begin{align}
    \dot{h}_i^p = &\;(\xs{i} - \traj(t))^\top(\Vtraj(t) - \us{i}), \\
\ddot{h}_i^p = &\;(\xs{i} - \traj(t))^\top(\Atraj(t) - \as{i}) - \norm{\us{i} - \Vtraj(t)}^2 .
\end{align}
Note that our control input appears in $\ddot{h}_i^p$ in $\as{i}$, and thus the CBF is of degree 2.
To enforce the forward invariance of $\mathcal{P}_i$, the constraint $h_i^p$ is restricted as 
\begin{align}
    v_i^p = \dot{h}_i^p + p_1 h_i^p\geq 0
\end{align}
where $v_i^p$ does not contain any control inputs. Further differentiating the above CBF yields
\begin{align}
    \omega_i^p &= \dot{v}_i^p + p_2 v_i^p = \ddot{h}_i^p + \alpha\dot{h}_i^p + \beta h_i^p \geq 0 
    \label{shep_herding_CBF}
\end{align}
where $\alpha = p_1 + p_2$ and $\beta = p_1 p_2$.
$w_i^p$ can be written to make the control input $\ud{}$ explicit as follows
\small
\begin{align}\nonumber
 -\sum_{j\in \mathcal{D}} (\xs{i} - \traj(t))^\top\frac{\partial \vect{f}_i}{\partial \xd{j}}\ud{j} \leq 
\frac{\beta}{2}(R^d-r)^2 -  \norm{\us{i} - \Vtraj(t)}^2 + \nonumber\\
    (\xs{i} - \traj(t))^\top \left(\Atraj(t) + \alpha ( \Vtraj(t) - \us{i}) + \frac{\beta}{2} (\traj(t) - \xs{i})\right)  + \nonumber\\
    (\xs{i} - \traj(t))^\top \left( -\sum_{j\in \mathcal{D}}\frac{\partial \vect{f}_i}{\partial \xd{j}} \us{i} - \sum_{\rev{k}\in \mathcal{S}/i}\frac{\partial \vect{f}_i}{\partial \xs{\rev{k}}} (\us{\rev{k}} - \us{i})  \right).
 \end{align}
\normalsize
Equation \eqref{shep_herding_CBF} can be written as $\vect{A}_i^p \ud{} \leq b_i^p$ where $\vect{A}_i^p  \rev{\in} \nR{1 \times 2 m}$ for each sheep. We concatenate these constraints such that $\vect{A}^p = [\vect{A}_1^{p,{\top}}, \dots, \vect{A}_n^{p,{\top}}]^{\top}$ and $\vect{b}^p = [b_1^p , \dots, b_n^p]^{\top}$. As such, the sheep herding CBF for all sheep can be applied by the following constraint on the dog velocities
\begin{align}\label{sheep_herding_CBF_all}
    \vect{A}^p \ud{} \leq \vect{b}^p.
\end{align}
%It can be noted that \eqref{sheep_herding_CBF_all} is a function of the different agents positions and velocities, the trajectory and the corresponding desired radius, in addition to their first and second derivatives.
%
\subsection{Obstacle avoidance CBF}
%
%Let the existence of $L$ obstacles, where each obstacle $l$ is denoted by its boundary $\partial B_l$.
It is desired that each dog avoids colliding with the boundary $\partial B_l$ of the $l$-th obstacle. %Note that we assume that any obstacle with a possible collision would have already been scanned by the corresponding dog.
Assuming that each dog starts its position outside the boundary of all obstacles, the desired constraint can be defined for the $j$-th dog 
\begin{align}
    \mathcal{L}_l^j &= \{\xd{j} \in \nR{2} | \norm{\xd{\rev{j}} -b^*}^2 \geq R_{\circ}^2\}, \\
    b^* &= \argmin_{b\in\partial B_l} \norm{\xd{\rev{j}} -b}^2
\end{align}
where $b\in\partial B_l \subset \nR{2}$, $b^*$ is the closest point on the obstacle boundary to the dog's current position, and $R_{\circ}$ denotes the required safety distance between a dog and an obstacle. 

To render the set $\mathcal{L}_l^j$ forward invariant, let the CBF
\begin{align}
    h_j^{\circ} (l) =  \frac{1}{2}(\norm{\xd{\rev{j}} -b^*}^2 - R_{\circ}^2) \geq 0
\end{align}
where its derivative is given by
\begin{align}
    \dot{h}_j^{\circ} (l) = (\xd{\rev{j}} -b^*)^\top\ud{j}.
\end{align}
The $j$-th dog's velocity is constrained as follows
\begin{align}
    \dot{h}_j^{\circ} (l) + \lambda h_j^{\circ} (l) \geq 0 \\
    -(\xd{\rev{j}} -b^*)^\top\ud{j} \leq  \frac{\lambda}{2}(\norm{\xd{\rev{j}} -b^*}^2 - R_{\circ}^2)
\end{align}
where $\lambda$ is the obstacle avoidance CBF gain. Let
%Let us define $\vect{A}_{j,l}^{\circ}$ and $b^{\circ}_{j,l}$ such that 
\begin{align}
    \vect{A}_{j,l}^{\circ} &\stackrel{\triangle}{=} \begin{bmatrix}
        \ldots (b^* - \xd{\rev{j}})^{\top} \ldots
    \end{bmatrix} \in \nR{1\times 2m},\\
        b^{\circ}_{j,l} &\stackrel{\triangle}{=}  \frac{\lambda}{2}(\norm{\xd{\rev{j}} -b^*}^2 - R_{\circ}^2)
\end{align}
where $\vect{A}_{j,l}^{\circ}$ is equal to  $(b^* - \xd{\rev{j}})^{\top}$ in the columns corresponding to $(2(j-1)+1, 2j)$, and zero otherwise.
The obstacle avoidance constraint %between the $j$-th dog and the $l$-th obstacle 
can be written as $\vect{A}_{j,l}^{\circ} \ud{} \leq b^{\circ}_{j,l}$. Concatenate the obstacle avoidance constraints for all dogs and all obstacles such that ${\vect{A}^{\circ} = [\vect{A}_{1,1}^{{\circ},\top}, \cdots, \vect{A}_{m,L}^{{\circ},\top} ]^{\top} \in \nR{m L \times  2m }}$ and $\vect{b}^{\circ} = [ b^{\circ}_{1,1}, \cdots, b^{\circ}_{m,L}]^{\top} \in \nR{m L }$. Using these entities and grouping all inequalities,  the dog obstacle avoidance constraint among all dogs and all obstacles is
\begin{align}\label{eq:full_dog_obstalces}
    \vect{A}^{\circ} \ud{} \leq \vect{b}^{\circ}.
\end{align}
\subsection{Agent collision avoidance CBF}
Inter-sheep and dog-sheep collisions are ensured due to the sheep dynamics described in~\eqref{eq:sheep_dynamics}. However, it is required to ensure dogs do not collide with each other while herding the sheep. Assume that dogs' positions at $t_0$ do not coincide, and  represent the collision avoidance constraint by the following set for the $k$-th and $j$-th dogs
\begin{align}
    \mathcal{A}_\rev{k}^j = \{\xd{\rev{k}},\xd{j} \in \nR{2} | \norm{\xd{\rev{k}} -\xd{j}}^2 \geq R_a^2\ \}
\end{align}
where $R_a$ denotes the required safety distance between dogs. To ensure the forward invariance of the above set, let the CBF%us define the following CBF between dogs $\rev{k}$ and $j$
\begin{align}
    h_{\rev{k},j}^a = \frac{1}{2}(\norm{\xd{\rev{k}} - \xd{j}}^2 - R_a^2) \geq 0.
\end{align}
It can be proven $h_{\rev{k},j}^a = h_{j,\rev{k}}^a$, while the CBF's derivative is
\begin{align}
    \dot{h}_{\rev{k},j}^a = (\xd{\rev{k}} - \xd{j})^\top(\ud{\rev{k}} - \ud{j}).
\end{align}
To ensure that no dog collisions occur, dog velocities should be chosen as
\begin{align}
     \dot{h}_{\rev{k},j}^a + \gamma {h}_{\rev{k},j}^a &\geq 0, \\
     (\xd{j} - \xd{\rev{k}})^\top(\ud{\rev{k}} - \ud{j}) &\leq  \frac{\gamma}{2}(\norm{\xd{\rev{k}} - \xd{j}}^2 - R_a^2)
\end{align}
where $\gamma$ is the inter-agent collision avoidance CBF gain.
To avoid collision between all dogs, it is required to apply the above constraints $\forall (\rev{k},j),~\rev{k} \neq j \in \mathcal{D}$.
Let %us define $\vect{A}_{\rev{k},j}^a$ and $b^a_{\rev{k},j}$ such that 
\begin{align}
    \vect{A}_{\rev{k},j}^a &\stackrel{\triangle}{=} \begin{bmatrix}
        \ldots \underbrace{(\xd{j} - \xd{\rev{k}})^\top}_{(\rev{k}-1)2+1: 2 \rev{k}} \ldots  \underbrace{(\xd{\rev{k}} - \xd{j})^{\top}}_{(j-1)2+1: 2j} \ldots
    \end{bmatrix},\\
        b^a_{\rev{k},j} &\stackrel{\triangle}{=}   \frac{\gamma}{2}(\norm{\xd{\rev{k}} - \xd{j}}^2 - R_a^2).
\end{align} 
Concatenating the dog avoidance constraints and using $\vect{A}^a = [\vect{A}_{1,2}^{a,{\top}},\ldots,\vect{A}_{m-1,m}^{a,{\top}}]^{\top}\in \nR{m(m-1)/2 \times 2m}$ and ${\vect{b}^a = [b^a_{1,2},\dots,b^a_{m-1,m}]^\top \in \nR{m(m-1)/2 }}$,
the dogs can avoid collision between each other by satisfying
\begin{align}\label{eq:full_dog_repulsion}
    \vect{A}^a \ud{} \leq \vect{b}^a.
\end{align}
\subsection{Objective Function Formulation}
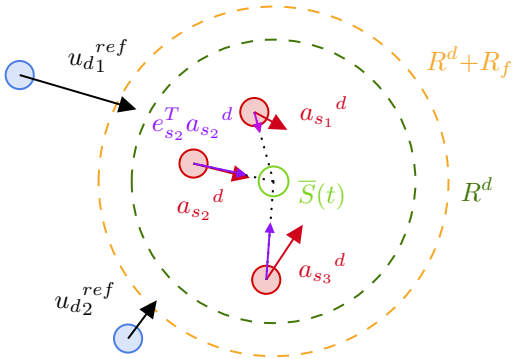
\begin{figure}[htbp]
    \centering
\tikzset{every picture/.style={line width=0.75pt}} %set default line width to 0.75pt        

\begin{tikzpicture}[x=0.75pt,y=0.75pt,yscale=-1,xscale=1]
%Shape: Circle [id:dp4679327946137608] 
\draw  [color={rgb, 255:red, 65; green, 117; blue, 5 }  ,draw opacity=1 ][dash pattern={on 4.5pt off 4.5pt}] (187.06,142.1) .. controls (187.06,102.59) and (219.09,70.56) .. (258.6,70.56) .. controls (298.11,70.56) and (330.14,102.59) .. (330.14,142.1) .. controls (330.14,181.61) and (298.11,213.64) .. (258.6,213.64) .. controls (219.09,213.64) and (187.06,181.61) .. (187.06,142.1) -- cycle ;
%Shape: Circle [id:dp09429996044977074] 
\draw  [color={rgb, 255:red, 245; green, 166; blue, 35 }  ,draw opacity=1 ][dash pattern={on 4.5pt off 4.5pt}] (170.37,142.1) .. controls (170.37,93.37) and (209.87,53.87) .. (258.6,53.87) .. controls (307.33,53.87) and (346.83,93.37) .. (346.83,142.1) .. controls (346.83,190.83) and (307.33,230.33) .. (258.6,230.33) .. controls (209.87,230.33) and (170.37,190.83) .. (170.37,142.1) -- cycle ;
%Shape: Circle [id:dp6444518319704515] 
\draw  [color={rgb, 255:red, 74; green, 125; blue, 226 }  ,draw opacity=1 ][fill={rgb, 255:red, 74; green, 125; blue, 226 }  ,fill opacity=0.2 ] (123.47,88.43) .. controls (123.47,84.51) and (126.65,81.33) .. (130.57,81.33) .. controls (134.49,81.33) and (137.67,84.51) .. (137.67,88.43) .. controls (137.67,92.35) and (134.49,95.53) .. (130.57,95.53) .. controls (126.65,95.53) and (123.47,92.35) .. (123.47,88.43) -- cycle ;
%Straight Lines [id:da09982188293988026] 
\draw    (130.57,88.43) -- (186.59,104.89) ;
\draw [shift={(189.47,105.73)}, rotate = 196.37] [fill={rgb, 255:red, 0; green, 0; blue, 0 }  ][line width=0.08]  [draw opacity=0] (8.93,-4.29) -- (0,0) -- (8.93,4.29) -- cycle    ;
%Shape: Circle [id:dp7708573900773712] 
\draw  [color={rgb, 255:red, 126; green, 211; blue, 33 }  ,draw opacity=1 ] (251.08,142.1) .. controls (251.08,137.95) and (254.45,134.58) .. (258.6,134.58) .. controls (262.75,134.58) and (266.12,137.95) .. (266.12,142.1) .. controls (266.12,146.25) and (262.75,149.62) .. (258.6,149.62) .. controls (254.45,149.62) and (251.08,146.25) .. (251.08,142.1) -- cycle ;
%Shape: Circle [id:dp009890421445151842] 
\draw  [color={rgb, 255:red, 74; green, 125; blue, 226 }  ,draw opacity=1 ][fill={rgb, 255:red, 74; green, 125; blue, 226 }  ,fill opacity=0.2 ] (178.13,221.43) .. controls (178.13,217.51) and (181.31,214.33) .. (185.23,214.33) .. controls (189.15,214.33) and (192.33,217.51) .. (192.33,221.43) .. controls (192.33,225.35) and (189.15,228.53) .. (185.23,228.53) .. controls (181.31,228.53) and (178.13,225.35) .. (178.13,221.43) -- cycle ;
%Straight Lines [id:da292360133038043] 
\draw    (185.23,221.43) -- (197.92,204.5) ;
\draw [shift={(199.72,202.1)}, rotate = 126.85] [fill={rgb, 255:red, 0; green, 0; blue, 0 }  ][line width=0.08]  [draw opacity=0] (8.93,-4.29) -- (0,0) -- (8.93,4.29) -- cycle    ;
%Shape: Circle [id:dp6437996535098112] 
\draw  [color={rgb, 255:red, 208; green, 2; blue, 2 }  ,draw opacity=1 ][fill={rgb, 255:red, 208; green, 2; blue, 2 }  ,fill opacity=0.2 ] (241.8,107.1) .. controls (241.8,103.18) and (244.98,100) .. (248.9,100) .. controls (252.82,100) and (256,103.18) .. (256,107.1) .. controls (256,111.02) and (252.82,114.2) .. (248.9,114.2) .. controls (244.98,114.2) and (241.8,111.02) .. (241.8,107.1) -- cycle ;
%Shape: Circle [id:dp4835853081974104] 
\draw  [color={rgb, 255:red, 208; green, 2; blue, 2 }  ,draw opacity=1 ][fill={rgb, 255:red, 208; green, 2; blue, 2 }  ,fill opacity=0.2 ] (211.13,133.1) .. controls (211.13,129.18) and (214.31,126) .. (218.23,126) .. controls (222.15,126) and (225.33,129.18) .. (225.33,133.1) .. controls (225.33,137.02) and (222.15,140.2) .. (218.23,140.2) .. controls (214.31,140.2) and (211.13,137.02) .. (211.13,133.1) -- cycle ;
%Shape: Circle [id:dp08744833315608203] 
\draw  [color={rgb, 255:red, 208; green, 2; blue, 2 }  ,draw opacity=1 ][fill={rgb, 255:red, 208; green, 2; blue, 2 }  ,fill opacity=0.2 ] (247.8,191.77) .. controls (247.8,187.85) and (250.98,184.67) .. (254.9,184.67) .. controls (258.82,184.67) and (262,187.85) .. (262,191.77) .. controls (262,195.69) and (258.82,198.87) .. (254.9,198.87) .. controls (250.98,198.87) and (247.8,195.69) .. (247.8,191.77) -- cycle ;
%Straight Lines [id:da08933967139816867] 
\draw [color={rgb, 255:red, 208; green, 2; blue, 27 }  ,draw opacity=1 ]   (248.9,107.1) -- (262.93,114.7) ;
\draw [shift={(265.57,116.13)}, rotate = 208.46] [fill={rgb, 255:red, 208; green, 2; blue, 27 }  ,fill opacity=1 ][line width=0.08]  [draw opacity=0] (8.93,-4.29) -- (0,0) -- (8.93,4.29) -- cycle    ;
%Straight Lines [id:da17539634385903025] 
\draw [color={rgb, 255:red, 208; green, 2; blue, 27 }  ,draw opacity=1 ]   (218.23,133.1) -- (243.79,139.56) ;
\draw [shift={(246.7,140.3)}, rotate = 194.19] [fill={rgb, 255:red, 208; green, 2; blue, 27 }  ,fill opacity=1 ][line width=0.08]  [draw opacity=0] (8.93,-4.29) -- (0,0) -- (8.93,4.29) -- cycle    ;
%Straight Lines [id:da317982962063023] 
\draw [color={rgb, 255:red, 208; green, 2; blue, 27 }  ,draw opacity=1 ]   (254.9,191.77) -- (271.71,166.47) ;
\draw [shift={(273.37,163.97)}, rotate = 123.59] [fill={rgb, 255:red, 208; green, 2; blue, 27 }  ,fill opacity=1 ][line width=0.08]  [draw opacity=0] (8.93,-4.29) -- (0,0) -- (8.93,4.29) -- cycle    ;
%Straight Lines [id:da28645863609010425] 
\draw  [dash pattern={on 0.84pt off 2.51pt}]  (258.6,142.1) -- (254.9,191.77) ;
%Straight Lines [id:da7602240865454135] 
\draw  [dash pattern={on 0.84pt off 2.51pt}]  (218.23,133.1) -- (258.6,142.1) ;
%Straight Lines [id:da5894379751671024] 
\draw  [dash pattern={on 0.84pt off 2.51pt}]  (248.9,107.1) -- (258.6,142.1) ;
%Straight Lines [id:da3991806741323525] 
\draw [color={rgb, 255:red, 189; green, 16; blue, 224 }  ,draw opacity=1 ]   (248.9,107.1) -- (251.01,115.02) ;
\draw [shift={(251.79,117.92)}, rotate = 255.06] [fill={rgb, 255:red, 189; green, 16; blue, 224 }  ,fill opacity=1 ][line width=0.08]  [draw opacity=0] (5.36,-2.57) -- (0,0) -- (5.36,2.57) -- cycle    ;
%Straight Lines [id:da9490733156632996] 
\draw [color={rgb, 255:red, 144; green, 19; blue, 254 }  ,draw opacity=1 ]   (218.23,133.1) -- (242.32,138.23) ;
\draw [shift={(245.25,138.85)}, rotate = 192.02] [fill={rgb, 255:red, 144; green, 19; blue, 254 }  ,fill opacity=1 ][line width=0.08]  [draw opacity=0] (5.36,-2.57) -- (0,0) -- (5.36,2.57) -- cycle    ;
%Straight Lines [id:da015196274568332502] 
\draw [color={rgb, 255:red, 144; green, 19; blue, 254 }  ,draw opacity=1 ]   (254.9,191.77) -- (256.77,165.69) ;
\draw [shift={(256.99,162.69)}, rotate = 94.11] [fill={rgb, 255:red, 144; green, 19; blue, 254 }  ,fill opacity=1 ][line width=0.08]  [draw opacity=0] (5.36,-2.57) -- (0,0) -- (5.36,2.57) -- cycle    ;

% Text Node
\draw (351.6,138.27) node [anchor=north west][inner sep=0.75pt]   [align=left] {$\displaystyle \textcolor[rgb]{0.25,0.46,0.02}{R^d}$};
% Text Node
\draw (334.13,72.27) node [anchor=north west][inner sep=0.75pt]   [align=left] {$\displaystyle \textcolor[rgb]{0.96,0.65,0.14}{R^d}\textcolor[rgb]{0.96,0.65,0.14}{+R}\textcolor[rgb]{0.96,0.65,0.14}{_{f}}$};
% Text Node
\draw (153.33,69) node [anchor=north west][inner sep=0.75pt]   [align=left] {$\displaystyle u_{d}{}_{1}^{ref}$};
% Text Node
\draw (269.52,139.71) node [anchor=north west][inner sep=0.75pt]   [align=left] {$\displaystyle \textcolor[rgb]{0.49,0.83,0.13}{\overline{S}}\textcolor[rgb]{0.49,0.83,0.13}{(}\textcolor[rgb]{0.49,0.83,0.13}{t}\textcolor[rgb]{0.49,0.83,0.13}{)}$};
% Text Node
\draw (146.67,191.6) node [anchor=north west][inner sep=0.75pt]   [align=left] {$\displaystyle u_{d}{}_{2}^{ref}$};
% Text Node
\draw (270.33,97.67) node [anchor=north west][inner sep=0.75pt]   [align=left] {$\displaystyle \textcolor[rgb]{0.82,0.01,0.11}{a}\textcolor[rgb]{0.82,0.01,0.11}{_{s_1}{}}\textcolor[rgb]{0.82,0.01,0.11}{^{d}}$};
% Text Node
\draw (208.33,144.67) node [anchor=north west][inner sep=0.75pt]   [align=left] {$\displaystyle \textcolor[rgb]{0.82,0.01,0.11}{a}\textcolor[rgb]{0.82,0.01,0.11}{_{s_2}{}}\textcolor[rgb]{0.82,0.01,0.11}{^{d}}$};
% Text Node
\draw (269.67,176.67) node [anchor=north west][inner sep=0.75pt]   [align=left] {$\displaystyle \textcolor[rgb]{0.82,0.01,0.11}{a}\textcolor[rgb]{0.82,0.01,0.11}{_{s_3}{}}\textcolor[rgb]{0.82,0.01,0.11}{^{d}}$};
% Text Node
\draw (195.72,103.31) node [anchor=north west][inner sep=0.75pt]  [color={rgb, 255:red, 144; green, 19; blue, 254 }  ,opacity=1 ] [align=left] {\textcolor[rgb]{0.56,0.07,1}{$\displaystyle e_{s_2}^{T}$}\textcolor[rgb]{0.56,0.07,1}{$\displaystyle \textcolor[rgb]{0.56,0.07,1}{a}\textcolor[rgb]{0.56,0.07,1}{_{s_2}{}}\textcolor[rgb]{0.56,0.07,1}{^{d}}$}};
\end{tikzpicture}
    \caption{Objective function components.}
    \label{fig:objective function}
\end{figure}

The aforementioned constraints ensure the desired sheep motion with any objective function, for example minimizing norm of dog actions. However, due to the bounded dog velocities, an objective function is employed to ensure that the sheep's accelerations can be modified within the dogs' action limits.
To understand this phenomenon, let $\as{i}^j$ be the acceleration the $j$-th dog applies on the $i$-th sheep;  $\as{i}^d$ be the sum of dogs' applied accelerations on the $i$-th sheep; the resultant acceleration can be seen in \fig\ref{fig:objective function} where $e_{s_i} =\traj(t) - \xs{i}$ .

This acceleration-term can be written as follows
\begin{align}
    \as{i}^j = \frac{\partial \vect{f}_i}{\partial \xd{\rev{j}}} \ud{j}
\end{align}
where  $\frac{\partial f_i}{\partial  \xd{j}} \rightarrow 0$ as the dogs' distance from the sheep gets larger. Consequently,  to apply any acceleration by the $j$-th dog on the $i$-th sheep, $\norm{\ud{j}} \rightarrow \infty$ as $\xd{j} \rightarrow \infty$. Thus, it becomes imperative for dogs to remain in proximity of all sheep. The proximity conditions $\frac{\partial \vect{f}_i}{\partial \xd{\rev{j}}}$ so that the dogs can modify the sheep' accelerations within the dogs' velocity limits. Define a reference velocity that keeps the dogs within a distance $R_f$ from the sphere containing the sheep
\begin{align}
     \ud{j}^{ref} &= k_f \;\text{relu}(\norm{\vect{e}_{dj}} - (R^d+R_f)) \frac{\vect{e}_{dj}}{\norm{\vect{e}_{dj}}},\\
     \vect{e}_{dj} &= \traj(t) - \xd{j}
\end{align}
where $k_f$ is a tunable control gain and  $\text{relu}(\cdot)$ is the rectified linear unit. The above velocity will bring the dogs to a distance $R^d + R_f$ from the current trajectory $\traj(t)$ while having no effect if the dogs are within the defined perimeter  as illustrated in \fig\ref{fig:objective function}. The reference velocities for the different dogs can be concatenated in a column vector $\ud{}^{ref} \in \nR{2m}$.

Furthermore, it is noted that a sheep's herding CBF constraint becomes active when the sheep gets closer to the edge of the protected region. As such, if a subgroup of sheep are close to the edge while others are far away, the dogs will attempt to herd only sheep closer to the edge while ignoring the resultant force applied on the other sheep. While this phenomenon can be avoided by fine tuning the CBF gains, the following minimization of a potential objective function alleviates its effect
\begin{align}
    \min_{\ud{}} & \lVert(\sum_{j\in \mathcal{D}} (\xs{i} - \traj(t))^{\top}\frac{\partial \vect{f}_i}{\partial \xd{j}}\ud{j} -\ \nonumber\\
     & \sum_{j\in \mathcal{D}} (\xs{k} - \traj(t))^{\top}\frac{\partial \vect{f}_k}{\partial \xd{j}}\ud{j}
\rVert^2~\forall (i,k), i\neq k \in \mathcal{S}.
\end{align}
This minimizes the difference between the dogs' applied acceleration on the $i$-th and $k$-th sheep, each projected on the vector between the sheep's corresponding positions and  $\traj(t)$. This minimization 
should be applied for each pair of sheep; the entities being minimized are illustrated in \fig\ref{fig:objective function}.
Conveniently, the above minimization between all pairs of sheep is
\begin{align}
\min_{\ud{}}\ud{}^\top\vect{A}^{p,\top}\vect{C}^\top \vect{C}\vect{A}^{p}\ud{}
\end{align}
where $\vect{C} \in \nR{n\times n}$ is a matrix with zeros in the first row, ones along the diagonal for the other rows, and -1 on the left columns of each 1 entry.
%For example, for $n=4$, 
%\begin{align}
%   \vect{C}= \begin{bmatrix}
%       0 & 0 & 0 & 0\\
%      -1 & 1 & 0 & 0\\
%        0& -1 & 1 & 0 \\
%       0 & 0 &-1 & 1 
%    \end{bmatrix}
%\end{align}
%The objective function minimizes the acceleration applied on each sheep by the dogs, projected on their corresponding distance to the center of the trajectory.

Hence, the complete control function subject to the detailed constraints can be computed through the following Quadratic Programming optimization
\begin{align}\label{eq:full_optimization}
    \ud{}^* = \argmin_{\vect{\mu}}& \vect{\mu}^\top\vect{A}^{p,\top}\vect{C}^\top \vect{C}\vect{A}^{p}\vect{\mu} +  \ud{}^{ref}\vect{\mu}\\\nonumber
    \mbox{subject to} \;\; &\vect{A}^p \vect{\mu} \leq \vect{b}^p\\\nonumber
    &\vect{A}^\circ \vect{\mu} \leq \vect{b}^\circ\\\nonumber
    &\vect{A}^a \vect{\mu} \leq \vect{b}^a\\\nonumber
    -\vect{1}\bar{u} &\leq \vect{\mu} \leq \vect{1}\bar{u}
\end{align}
where $\vect{1}$ is an all ones column vector.

%\paragraph{Existence of Solution}
The existence of a solution for the herding constraint \eqref{sheep_herding_CBF_all} when coupled with the other constraints is discussed hereafter.

To study the existence of solutions for \eqref{sheep_herding_CBF_all}, let the inequality be rewritten as
\begin{align}
     \vect{A}^p \ud{} \leq \vect{b}^p_s + \vect{b}^p_d 
\end{align}
where the vector $\vect{b}^p$ was split into a dynamic part $\vect{b}^p_d$ that contains all components related to $\Vtraj(t)$ and $\Atraj(t)$, and the remaining parts are contained in the static part $\vect{b}^p_s$.
\begin{prop}
Constraint \eqref{sheep_herding_CBF_all} is guaranteed to have a solution if $n=m$ and in the absence of the other constraints \eqref{eq:full_dog_obstalces}, \eqref{eq:full_dog_repulsion} and the dogs' velocity limits $\bar{u}$.
\end{prop}
\proof Similar to \cite{mohanty2022distributed}, we study the existence of a solution when each dog herds only a specific sheep. The existence of a solution when all dogs herd all sheep is guaranteed to exist as a consequence. We denote by index $i$ as the corresponding matrices related to the $i$-th sheep and dog.
The proof in this case can be established by ruling out the cases where a solution does not exist as follows
\begin{itemize}
    \item either when $\vect{A}^p_i = 0$ and $\vect{b}^p_i < 0$,
    \item or when $\vect{b}^p_i = -\infty$.
\end{itemize}

As shown in \cite{mohanty2022distributed}, 
$\vect{A}^p_i \neq 0$ as long as the distance between the $i$-th dog and sheep are finite, ruling out the first case.
In the second case, the authors in \cite{mohanty2022distributed} showed that 
$\vect{b}^p_{s,i}$ is finite as long as there exists an upper and lower bound on the distance between any pair of sheep, a lower bound on the distance between each dog and sheep, and an upper and lower bound on the dog velocities. Similarly, $\vect{b}^p_{d,i}$ is finite due to the above mentioned limits, in addition to the boundedness of $\Vtraj(t)$, $\Atraj(t) \; \forall t$; note that the necessary boundedness of the sheep velocities are guaranteed due to the above mentioned bounds. 
As such, in the case of $n=m$, the constraint \eqref{sheep_herding_CBF_all} always has a solution.

While the proof only holds for $n=m$ and in the absence of other constraints, it is along similar lines as the work in \cite{grover2022noncooperative,mohanty2022distributed} that the dogs are capable of maintaining the sheep within the desired region despite the additional constraints and when $n>m$.
In the case where the constraint from \eqref{sheep_herding_CBF_all} is not feasible concurrently with the other constraints, the herding constraint is allocated as a soft constraint, thus minimizing $\vect{A}^p\vect{\mu}-\vect{b}^p$
within the limits of the other constraints. %This situation arises for example when dogs are required to apply a velocity beyond their actuation limits.

\subsection{Overall Herding Approach}
The full shepherding approach is summarized below in Algorithm~\ref{alg:shepherding}, where $\mathcal{I}$ integrates the new position of an agent using a Runge-Kutta integrator, and $dt$ is a time step between two consecutive iterations. 
In this algorithm, a path and trajectory are computed as discussed in Sec.\ref{subsec:path}. The dogs then enforce the CBFs from~\eqref{eq:full_optimization} with the generated trajectory. The trajectory is updated each time the dogs receive a full depth scan of the environment, \ie the LiDAR sensor does a full rotation. If the current trajectory segment is no longer valid, and a new path has not yet been computed, a new trajectory is fitted to a new segment starting from the current position in the previously computed path.

\begin{algorithm}
\caption{Shepherding Algorithm}\label{alg:shepherding}
\begin{algorithmic}
\State {\bf Given:} $\traj_f$, $R^d$, $\overline{\Vtraj}$, $\overline{\Atraj}$ 
\State $\traj(t_0)$ = mean($\xs{i},~i = [1,\ldots,n])$
\State Initialize $\mathcal{M}$ as unoccupied
\State Generate path $\pathVar{}[k] \forall k\in[0,...,K]$~\cite{unlu2023uav}, and fit trajectory $\traj(t) \forall t \in [t_c, t_c+T]$
\While{$t\leq t_f$}%  $\traj(t) \neq \traj_f$}
\If {$\traj(t) \neq \traj_f$}
\If{$\pathVar{}[H] \neq \traj_f$ and new $\measj{j}(t)$ received}
\State $\mathcal{M} \leftarrow \text{UpdateOccupancyMap}(\mathcal{M} | \measj{j}(t), \xd{j}(t))$
\State Generate path $\pathVar{}[k] \forall k\in[0,...,K]$~\cite{unlu2023uav}, and fit trajectory $\traj(t) \forall t \in [t_c, t_c+T]$
%\State Generate trajectory $\traj(t,\eta^*)$ from Algorithm~\ref{alg:traj_generation}
\EndIf
\If{$t= t_c+T$}
\State Fit new trajectory segment $\traj(t) \forall t \in [t_c, t_c+T]$
\EndIf
\EndIf
\State $\ud{}^* $= argmin$_{\vect{\mu}}$ from \eqref{eq:full_optimization}
\State $\xd{} \gets \mathcal{I}(\xd{}, \ud{}, dt)$
\State $\us{i} = \vect{f}_i \, \forall i \in \mathcal{S}$ \eqref{eq:sheep_dynamics}
\State  $\xs{} \gets \mathcal{I}(\xs{}, \us{}, dt)$
\EndWhile
\end{algorithmic}
\end{algorithm}

%%%%%%%%%%%%%%%%%%%%%%%%%%%%%%%%%%%%%%%%%%%%%%%%%%%%%%%%%%%%%%%%%%%%%%
\section{Simulation Studies}\label{sec:results}
%%%%%%%%%%%%%%%%%%%%%%%%%%%%%%%%%%%%%%%%%%%%%%%%%%%%%%%%%%%%%%%%%%%%%%
The case of dogs steering a herd of sheep from an initial position to the final desired position is provided. Initially, simulations in an obstacle free environment are provided to demonstrate the performance of trajectory enforcement, followed by the case of two obstacle cluttered environments. %We provide multiple simulations in a 2D space, and one simulation in a 3D space. 
In all simulations, the number of sheep is larger than the number of dogs $(n > m)$.
%It should be noted that the provided videos demonstrate the step-by-step performance of the presented algorithm for each of the following cases.

The simulations were computed in Matlab and the integration was performed at $100 \,[\rm Hz]$. Each agent is modeled as a circle with a radius of $0.1\,[\rm m]$, corresponding to a small sheep/dog robot. 
The dynamic parameters are as follows: $k_s = 0.3$, $k_d = 0.15$, $R_s = 0.5\,[\rm m]$, $R_a = 0.2\,[\rm m]$. All mobile agents have equal velocity limits ${\bar{u} = \bar{v} = 0.4\,[\rm m]}$. The CBF gains are chosen such that they provide a smooth performance, with $p_1 = 5.2$ and $p_2 = 8.2$. 

It was noticed that the CBF algorithm maintains sheep within the protected region for a range of gains. However, the selection of gain values changes the dogs' behavior. Lower CBF-gains emphasize safety leading to more rapid dog responses to sheep movements. Conversely, higher CBF-gains reflect a more relaxed safety requirement, resulting in smoother dog motion.
The CBF algorithm maintains sheep inside the circle for a wide range of $R_d$; in our simulation, we automated the choice of $R_d$ such that $R_d = \overline{R}_s + r_s$, where $r_s = 0.35\,[\rm m]$ is a safety distance. In each simulation, $\overline{R}_s$ was estimated based on the herd's steady-state behavior observed at the beginning of the simulation, when the sheep had formed their steady-state topological cluster and while the dogs were distant enough to minimally disturb the formation.
%The value of $r_s$ was chosen to provide a smooth behavior of the sheep and dogs such that $r_s = 0.35\,[\rm m]$.  Note that $\overline{R}_s$ changes between simulations depending on the number of agents.

In terms of the mapping and trajectory planning, each dog was fitted with a $360^\circ$ LiDAR with a maximum range of $5\,[\rm m]$, taking measurements every $2^\circ$, and a circle scan frequency of $1\,[\rm Hz]$.
The kinematic constraints of the fitted trajectory were chosen such that $\overline{\Vtraj} = 0.48\,[\rm m/s]$ and $\overline{\Atraj} = 0.2\,[\rm m/s^2]$, and the trajectory segment fitting window was chosen such that $H=50$ grid points. %Finally, $t_f$ was chosen large enough to ensure that $\traj(t_f) = \traj_f$.

\subsection{Obstacle-Free Herding}
\begin{figure}[tbh]
     \centering
     %[trim={left bottom right top},clip]
     \includegraphics[trim={10 20 20 40},clip, width = 0.8\columnwidth]{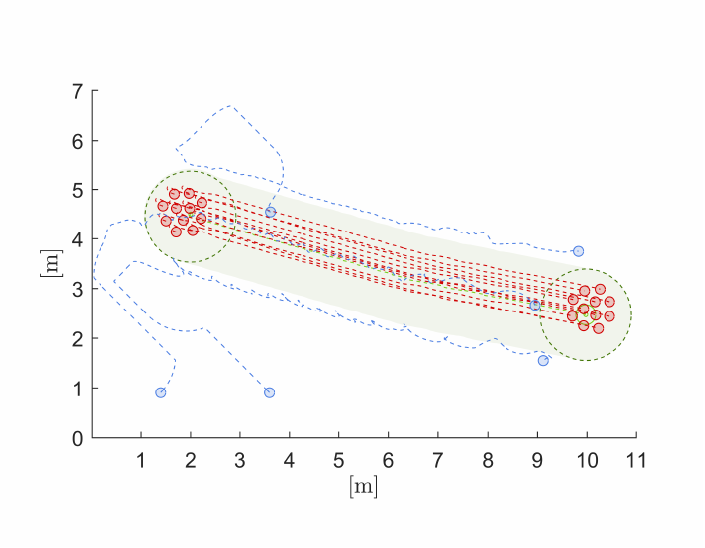}
     \caption{Herding in an obstacle-free environment.}
     \label{fig:2d_7_3_nobs}
 \end{figure}

Fig.~\ref{fig:2d_7_3_nobs} shows the herding performance in the absence of obstacles, having 11 sheep and 3 dogs. The herding trajectory is shown in dashed green line, while the green area around the trajectory shows the intersection of the safe regions along the trajectory. 

This scenario shows that while the algorithm was not designed to explicitly cage the sheep, the dogs form a loose single-sided cage around the sheep, while having a resultant acceleration in the direction of the trajectory.
After the cage is formed, the dogs push the sheep along the desired trajectory, and then place themselves equally distant from the sheep to halt their motion at the end of the trajectory.
Finally, from this simulation, it can be seen that the dogs and sheep motion is smooth along the trajectory.% in the absence of obstacles.
\subsection{Cluttered and Unknown Environment Herding}
%Figures~\ref{fig:2d_3_2} and \ref{fig:2d_7_3} show two simulations where 
In the following simulations, the dogs scan the environment to find a safe path to herd the sheep from their starting position to the final desired location.
\subsubsection{Small Herding-size in a Confined Area}
\begin{figure}[tbhp]
     \centering
     %[trim={left bottom right top},clip]
     \includegraphics[trim={20 00 30 20},clip, width = 0.9\columnwidth]{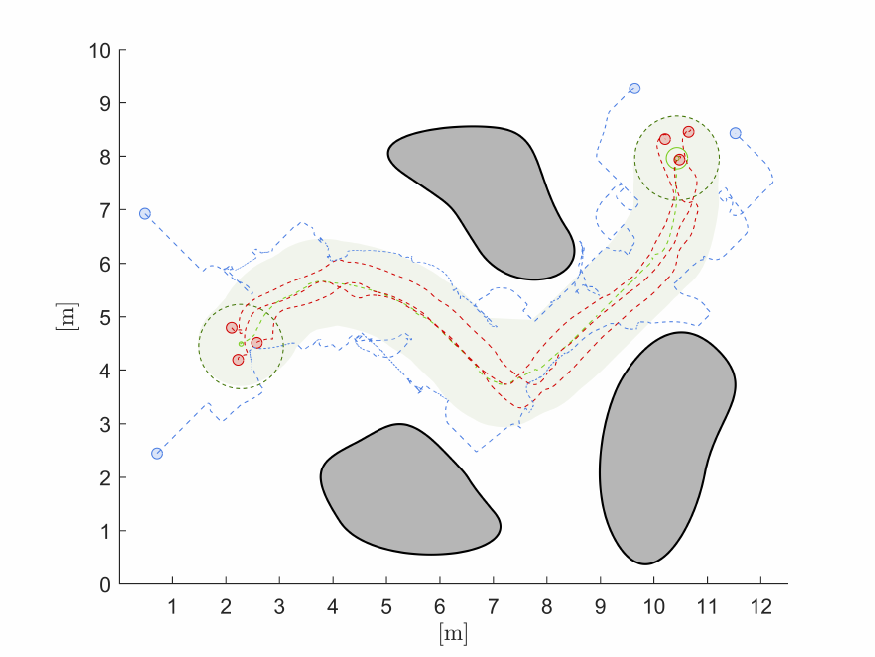}
     \caption{Herding in a small-size cluttered space.}
     \label{fig:2d_3_2}
 \end{figure}

\begin{figure}[tbh]
     \centering
     %[trim={left bottom right top},clip]
     \includegraphics[trim={0 0 0 0},clip, width = 0.92\columnwidth]{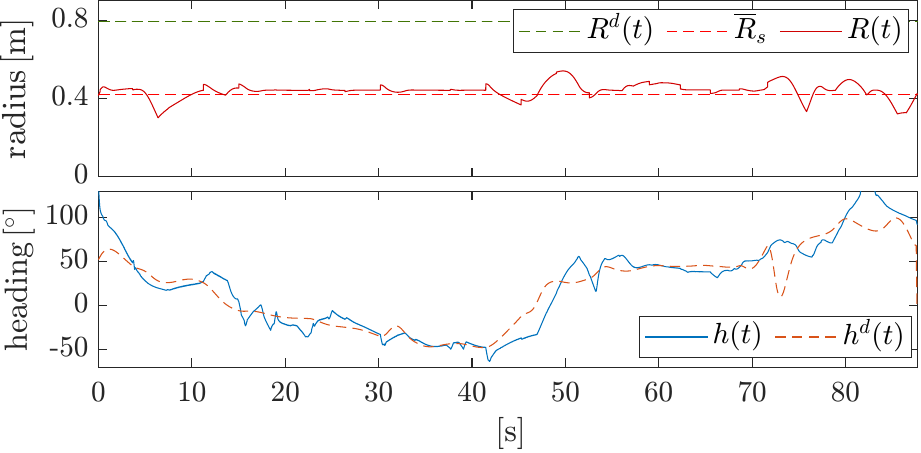}
     \caption{Collective sheep herd position and orientation (top). Desired and actual heading $h^d(t)$, $h(t)$ of the sheep swarm (bottom).}
     \label{fig:heading_analysis}
 \end{figure}

Fig.~\ref{fig:2d_3_2} shows 2 dogs herding 3 sheep in a small area. The radius of the minimum circle containing all sheep centered at $\traj(t)$, denoted by $R(t)$ and the mean heading of the sheep $h(t)$ during this herding simulation are shown in Fig.~\ref{fig:heading_analysis}. In the same figure, the desired circle bound $R^d$, minimum sheep radius $\overline{R}_s$, and desired sheep heading $h^d(t)$ are also displayed.

%The map in this simulation was purposefully bounded around the area shown in the figure to avoid taking the path around the obstacles. 
In this study the dogs were required to drastically change the orientation of the sheep motion. This direction change can be seen in Fig.~\ref{fig:heading_analysis} at $t = 42\,[\rm s]$, where the desired direction changes $80^\circ$ in $7 \,[\rm s]$. It can also be seen from Fig.~\ref{fig:heading_analysis} that the center of the sheep herd is kept safely away from the border of the safe region, where $R(t) < R^d ~\forall t$.  

\begin{figure}[tbhp]
     \centering
     %[trim={left bottom right top},clip]
     \includegraphics[trim={0 0 0 0},clip, width = 0.95\columnwidth]{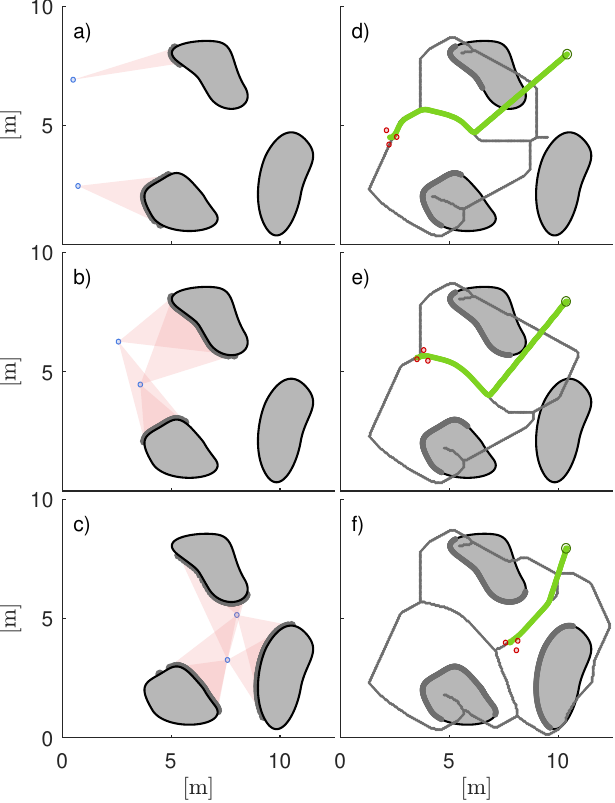}
     \caption{showing (a-c) LiDAR scans at three different time instances, highlighting the detected obstacles by the current scans and (d-e) corresponding cumulatively discovered obstacle frontiers, skeleton of the discovered map, and constructed path from $
     \traj(t)$ to $\traj_f$.}
     \label{fig:skeletal}
 \end{figure}

Fig. \ref{fig:skeletal} shows the environment mapping and the computed path for the above herding scenario with the LiDAR scans in Figures \ref{fig:skeletal}a-\ref{fig:skeletal}c. The corresponding frontier of the obstacles, computed skeleton about the obstacles, and computed path are shown in Figures \ref{fig:skeletal}d-\ref{fig:skeletal}e.
 
 \subsubsection{Herding in an Obstacle-Cluttered Environment}
\begin{figure}[tbhp]
     \centering
     %[trim={left bottom right top},clip]
     \includegraphics[trim={20 00 30 10},clip, width = 1.\columnwidth]{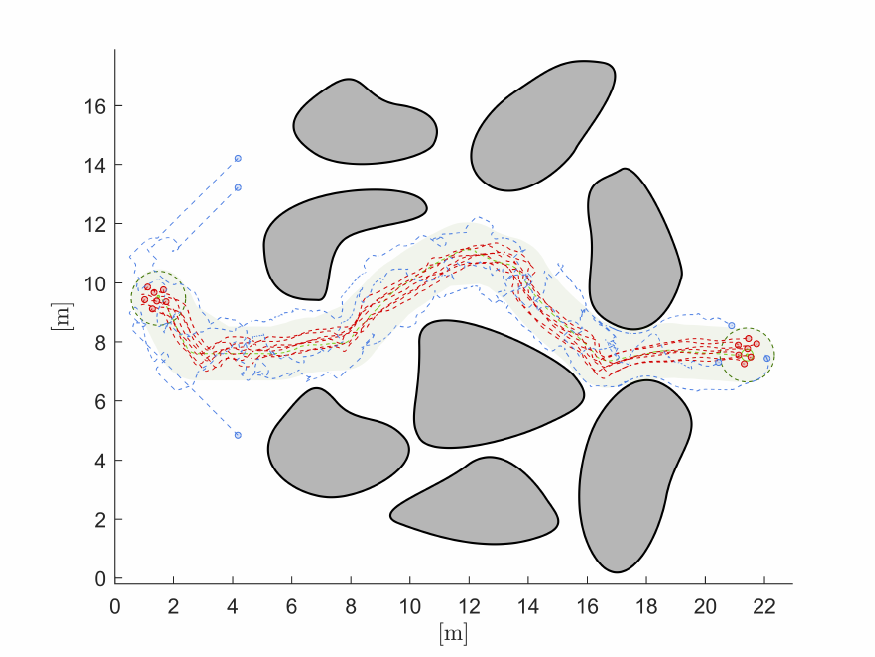}
     \caption{Herding in an obstacle-cluttered maze-like environment.}
     \label{fig:2d_7_3}
 \end{figure}

Fig.~\ref{fig:2d_7_3} shows a herding simulation in a maze-like cluttered environment using 7 sheep and 3 dogs. In this simulation, the dogs have to progressively explore the environment and re-plan the trajectory until the target has been reached. The environment in Fig.~\ref{fig:2d_7_3} has multiple openings which are traversable by the classical skeletal approach; however, since openings smaller than $\overline{R}_s$ where marked as untraversable, the path planning algorithm avoided such routes and found the  safest route. 

%%%%%%%%%%%%%%%%%%%%%%%%%%%%%%%%%%%%%%%%%%%%%%%%%%%%%%%%%%%%%%%%%%%%%%
\section{CONCLUSIONS}\label{sec:conclusions}
This paper presents a novel method for robot-sheep herding using a small number of robot-dogs in a cluttered and unknown environment. The presented optimization based controller ensures the sheep herd follows the desired trajectory, computed from the  partial map generated by LiDARs (measuring distances to obstacles) using the skeletal approach.  The presented approach was demonstrated in multiple simulation studies showing the effective herding of the sheep in cluttered and unknown environments, and  safety of the computed trajectory.

\bibliographystyle{IEEEtran}
\bibliography{./bibAlias,./bibCustom}
\end{document}